\documentclass[final]{cvpr}

\usepackage{times}
\usepackage{epsfig}
\usepackage{graphicx}
\usepackage[utf8]{inputenc} 
\usepackage[T1]{fontenc}    
\usepackage{url}            
\usepackage{booktabs}       
\usepackage{amsfonts}       
\usepackage{nicefrac}       
\usepackage{microtype}      

\usepackage{amsmath}
\usepackage{bbm}

\usepackage{algorithm}
\usepackage{algorithmic}
\usepackage{subfigure}
\usepackage{epstopdf}
\usepackage{booktabs}
\usepackage{array}
\usepackage{multirow}
\usepackage{hhline}
\usepackage{makecell}
\usepackage{framed}
\usepackage{enumitem}

\usepackage{xcolor}

\usepackage{textcomp}

\usepackage[pagebackref=true,breaklinks=true,colorlinks,bookmarks=false]{hyperref}

\def\cvprPaperID{2610} 
\def\confYear{CVPR 2021}

\graphicspath{{images/}}

\makeatletter
\newcommand{\printfnsymbol}[1]{%
	\textsuperscript{\@fnsymbol{#1}}%
}
\makeatother

\begin{document}

\title{Robust Neural Routing Through Space Partitions for Camera Relocalization\\ in Dynamic Indoor Environments}

\author{
	Siyan Dong$^{1,2}$\thanks{Equal Contribution} \quad Qingnan Fan$^{3}$\printfnsymbol{1} \quad He Wang$^{3}$ \quad Ji Shi$^{4}$ \quad Li Yi$^{5}$ \\ \quad Thomas Funkhouser$^{5}$ \quad Baoquan Chen$^{4}$ \quad Leonidas Guibas$^{3}$\\
	$^1${Shandong University} \quad	$^2${AICFVE, Beijing Film Academy} \quad $^3${Stanford University} \\
	\quad $^4${Peking University} \quad $^5${Google Research}\\
	{\tt\small \{siyandong.3,fqnchina,baoquan.chen\}@gmail.com, hewang@stanford.edu, i@sjj118.com,}\\
	{\tt\small \{ericyi,tfunkhouser\}@google.com, guibas@cs.stanford.edu}
}

\maketitle

\begin{abstract}
   Localizing the camera in a known indoor environment is a key building block for scene mapping, robot navigation, AR, \textit{etc}. Recent advances estimate the camera pose via optimization over the 2D/3D-3D correspondences established between the coordinates in 2D/3D camera space and 3D world space. Such a mapping is estimated with either a convolution neural network or a decision tree using only the static input image sequence, which makes these approaches vulnerable to dynamic indoor environments that are quite common yet challenging in the real world. To address the aforementioned issues, in this paper, we propose a novel outlier-aware neural tree which bridges the two worlds, deep learning and decision tree approaches. It builds on three important blocks: (a) a hierarchical space partition over the indoor scene to construct the decision tree; (b) a neural routing function, implemented as a deep classification network, employed for better 3D scene understanding; and (c) an outlier rejection module used to filter out dynamic points during the hierarchical routing process. Our proposed algorithm is evaluated on the RIO-10 benchmark developed for camera relocalization in dynamic indoor environments. It achieves robust neural routing through space partitions and outperforms the state-of-the-art approaches by around 30\% on camera pose accuracy, while running comparably fast for evaluation.
\end{abstract}

\section{Introduction}

    
    The task of camera relocalization is to estimate the 6-DoF (Degree of Freedom) camera pose from a test frame with respect to a known environment. 
    It is of great importance for many computer vision and robotics applications, such as Simultaneously Localization and Mapping (SLAM), Augmented Reality (AR), and navigation, \textit{etc}. 
    One popular solution to camera relocalization is to make use of advanced hardware, \textit{e.g.}, LIDAR sensors, WIFI, Bluetooth or GPS. However, these approaches may suffer from bad weather for outdoor environments, and instability or blocked signal for indoor environments. Another popular solution replaces the above hardware with a RGB/RGB-D sensor that feeds only visual observation for camera relocalization, also known as visual relocalization, which is the focus of this paper.
    
    
    \begin{figure}[t]
		\centering
		\includegraphics[width=0.94\linewidth]{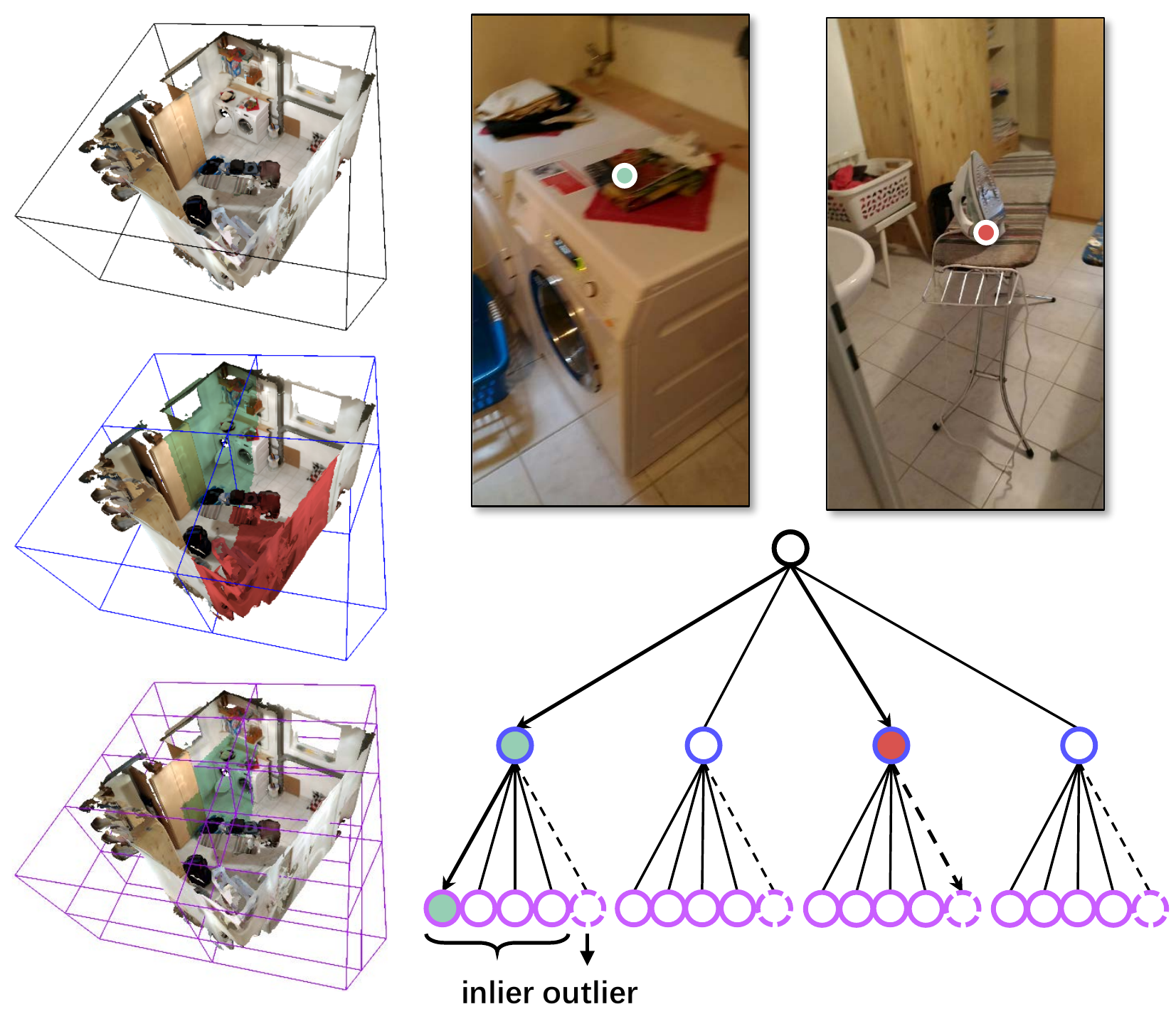}
		\caption{Demonstration of our algorithm. We build a hierarchical space partition over the entire scene environment to construct a 3-level 4-way neural tree. For the input static (green) or dynamic (red) points from a visual observation, our neural tree will route them into either inlier (solid line) or outlier (dashed line) categories. Only the points falling into the inlier category will be considered for camera pose estimation.}		
		\label{figure:teaser}
 		\vspace{-6mm}
	\end{figure}
	
    The problem of visual relocalization has been studied for decades, and recent advances \cite{cavallari2019real,li2019hierarchical} have reached around 100\% camera pose accuracy (5cm / 5$^\circ$) on the popular indoor scene benchmarks 7-scenes \cite{shotton2013scene} and 12-scenes \cite{valentin2016learning}. One type of successful approach in this regard is designed based on decision trees, which was firstly introduced into the camera relocalization field in \cite{shotton2013scene}, with many follow-ups \cite{massiceti2017random,meng2017backtracking,meng2018exploiting,cavallari2017fly,cavallari2019real}. They build a binary regression forest that takes a query image point sampled from the visual observation as input, and routes it into one leaf node via a hierarchical splitting strategy, which is simply implemented as color/depth comparison within the neighbourhood of the query point. The leaf node fits a density distribution over the 3D world coordinates from the training scene. Hence, by evaluating the decision tree with a test image, a 2D/3D-3D correspondence can be easily established between the input sample and regressed world coordinate for camera pose optimization.
    
    Although the aforementioned approaches are good at camera relocalization in static training environments, they tend to fail in dynamic test scenes, which are quite common yet challenging in real life. This is mainly due to the fact that the decision tree is constructed using only the static training image sequence so that, for any image point belonging to dynamic regions captured during evaluation, it is challenging to locate its correct correspondence in the leaf node. Recent studies \cite{wald2020beyond} have demonstrated that the decision tree based approaches achieve around 28\% camera pose accuracy (5cm 5$^\circ$), which is also the best among all the competitors, in their proposed RIO-10 benchmark developed for dynamic indoor scenes. This performance is far from being comparable to the ones in static indoor scenes.
	
	
	
	In order to tackle the challenges of camera relocalization in dynamic indoor environments, in this paper, we propose to learn an \textit{outlier-aware neural tree} to help establish point correspondences for accurate camera pose estimation focusing only on the confident static regions of the environment. Our algorithm inherits the general framework of decision trees, but mainly differs in the following aspects in order to obtain better generalization ability in dynamic test scenes.
	(a) \textbf{Hierarchical space partition.} We perform an explicit hierarchical spatial partition of the 3D scene in the world space to construct the decision tree. Then each split node in the decision tree not only performs a hard data partition selection, but in fact one which also corresponds to a physically-meaningful 3D geometric region.
	(b) \textbf{Neural routing function.} Given an input point sampled from the 2D visual observation, the split node needs to determine which divided sub-region in the world space to go. Such a classification task needs more contextual understanding of the 3D environment. Therefore, we propose a neural routing function, implemented as a deep classification network, for learning the splitting strategy. 
	(c) \textbf{Outlier rejection.} In order to deal with potential dynamic input points, we propose to consider these points as outliers and reject them during the hierarchical routing process in the decision tree. Specifically, the neural routing function learns to classify any input point from the dynamic region into the outlier category, stopping any further routing for that point. Once our proposed neural tree is fully trained, we follow the optimization and refinement steps in existing works \cite{cavallari2017fly,cavallari2019real} to calculate the final pose.
	
	We further train and test our proposed outlier-aware neural tree on the recent camera relocalization benchmark, RIO-10, which aims for dynamic indoor scenes. Experimental results demonstrate that our proposed algorithm outperforms the state-of-the-art localization approaches by at least 30\% on camera pose accuracy. More analysis shows that our algorithm is robust to various types of scene changes and successfully rejects most dynamic input samples during neural routing.
	
\section{Related Work}

    \subsection{Camera Relocalization}
    
    \textbf{Direct pose estimation.} 
    Approaches of this type aim for predicting the camera pose directly from the input image. One popular solution in this direction is image retrieval \cite{gee20126d,galvez2011real,glocker2014real,arandjelovic2016netvlad,jegou2010aggregating}. They approximate the camera pose of the query image by matching the most similar images in the dataset with known poses using low-level image features. 
    Instead of matching features, PoseNet \cite{kendall2015posenet} and many follow-ups \cite{kendall2017geometric,wang2019atloc,brahmbhatt2018geometry,walch2017image,sattler2019understanding} propose to use a convolution neural network to directly regress the 6D camera pose from an input image. 
    However, as mentioned in \cite{sattler2019understanding}, the performance of direct pose regression using CNNs is more similar to the one using image retrieval, and still lags behind the 3D structure-based approaches detailed below.
	
	\textbf{Indirect pose estimation.} 
	Approaches of this type find correspondences between camera and world coordinate points, and calculate the camera pose through optimization with RANSAC \cite{chum2003locally}. One common direction is to leverage the 2D-3D correspondences between traditional keypoints in the observed image and 3D scene map \cite{sattler2011fast,lim2012real,sattler2016large,sattler2016efficient}, followed by some recent works that deploy deep learning features \cite{sarlin2018leveraging,sarlin2019coarse,taira2018inloc,dusmanu2019d2} to replace the extracted poor descriptors.
	Another common direction to seek correspondences is scene coordinate regression.
	Shotton \textit{et al.} \cite{shotton2013scene} proposes to regress the 3D points in the world space from a query image point by training a decision tree, followed by many variants \cite{meng2017backtracking,meng2018exploiting,brachmann2016uncertainty,valentin2015exploiting}. 
	The other related works \cite{brachmann2017dsac,brachmann2018learning,brachmann2019neural,brachmann2019expert,li2018full,yang2019sanet,li2019hierarchical,massiceti2017random} in this direction leverage the deep convolutional neural network to regress the world coordinates from an input image, with a following pose optimization step.

	\begin{figure*}[t]
		\centering
		\includegraphics[width=0.94\linewidth]{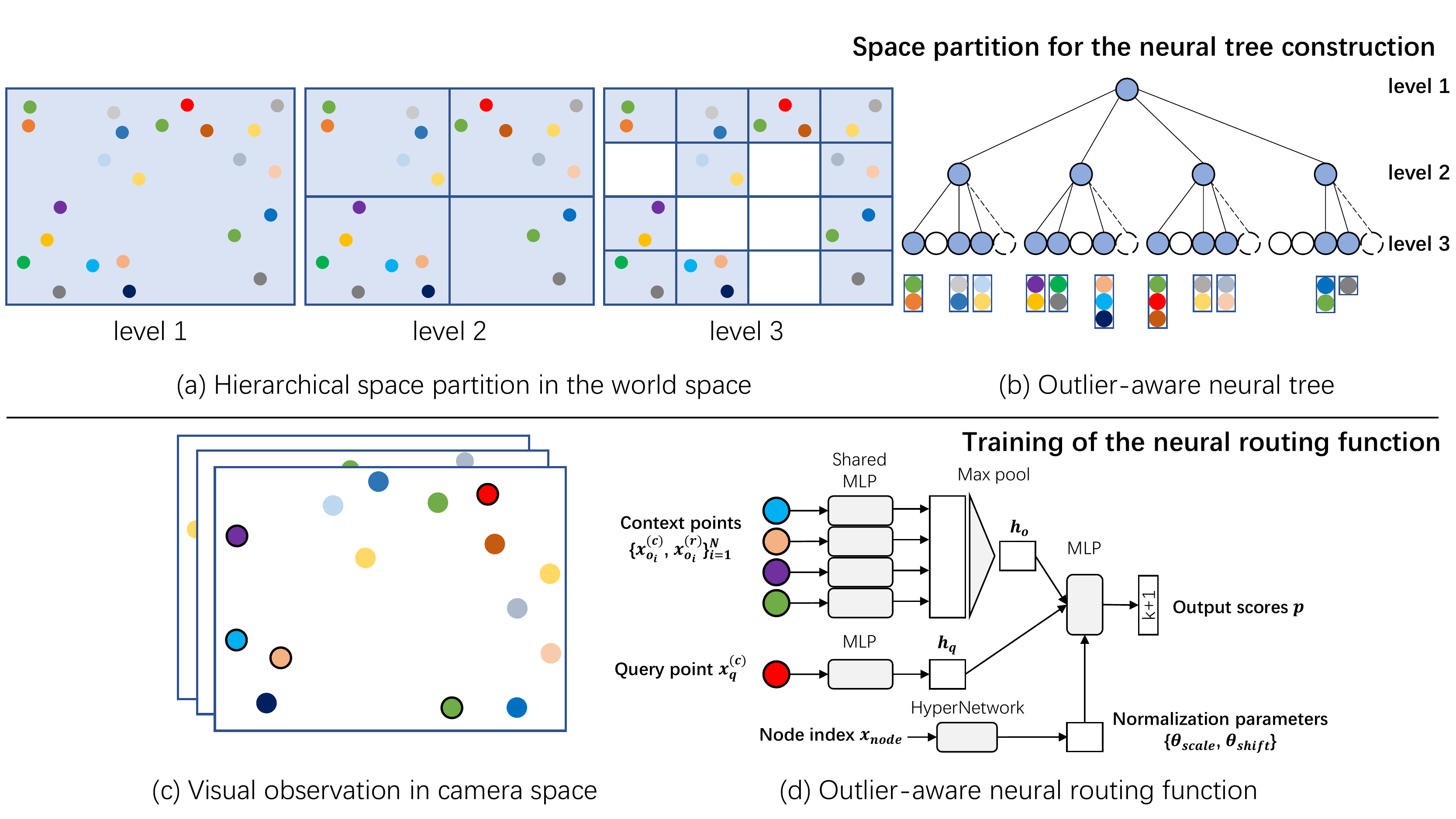}
		\caption{Illustration of our algorithm on the simple 2D case. Top: constructing a 3-level 4-way outlier-aware neural tree of a scene environment via hierarchical space partition. The dashed line and circle indicates the outlier category designed to reject the input dynamic points. Bottom: training an outlier-aware neural routing function for each split node in the neural tree.}		\label{figure:framework}
 		\vspace{-5mm}
	\end{figure*}
	
	\subsection{Decision Tree and Deep Learning.}
	
	Some recent efforts have been devoted to combining the two families of decision tree and deep learning techniques.
	The deep neural decision trees \cite{kontschieder2015deep} propose a principled, joint and global optimization of split and leaf node parameters, and hence enable end-to-end differentiable training of the whole decision tree. Shen \textit{et al.} \cite{shen2017label} presents label distribution learning tree to enable all the decision trees in a forest to be learned jointly. The variants of deep neural decision trees have been successfully applied for the task of human age estimation \cite{shen2018deep} and monocular depth estimation \cite{roy2016monocular}. Most of the aforementioned works formulate the last few fully connected layers in a classification neural network with the decision tree structure, and hence are significantly different from our algorithm.
	
	\section{NeuralRouting}
	\subsection{Overview}
	
	The input to our algorithm is a training sequence of <RGB-D image, camera pose> and a test frame for camera relocalization.
	Our algorithm can be separated into two steps, \textit{scene coordinate regression} and \textit{camera pose estimation}. The former step is conducted by learning a neural tree that takes a query point along with its neighbor context as input, and regresses its scene coordinate in the world space to build a 3D-3D correspondence, based on which, the latter step estimates the camera pose via iterative optimization followed by an optional ICP refinement. The neural tree is constructed via performing an explicit space partition in the scene environment, and learns to reject the dynamic points as outliers during the hierarchical routing process. In this way, our algorithm learns to build the 3D-3D correspondence only within the confident static region for accurate camera pose optimization. We firstly revisit the decision tree and its adaptation for camera relocalizatoin in Sec. \ref{sec:revisit}, and introduce our outlier-aware neural tree developed for relocalization in dynamic environments in Sec. \ref{sec:tree}. Finally, we describe the camera pose optimization and refinement details in Sec. \ref{sec:pose}.
    
    \vspace{-1mm}
    \subsection{Decision Tree for Camera Relocalization} \label{sec:revisit}
    \vspace{-1mm}
    
    Depending on whether the target is continuous or discrete, the decision tree can be used for either regression or classification tasks. A decision tree consists of a set of split nodes and leaf nodes. Each split node is assigned with a routing function, which learns the decision rules for the input sample partition, and each leaf node contains a probability density distribution fitted for the partitioned data. Given an input sample, inference starts from the root node and descends level-by-level until reaching the leaf node by evaluating the routing functions. A standard decision tree is binary, and employs greedy algorithms to learn the parameters at each split node to achieve locally-optimal hard data partition.
    
    For the task of camera relocalization, the decision tree \cite{cavallari2019real} is used to build the 3D knowledge of the known environment using the provided training sequence. Each split node takes a query point (RGB-D) sampled from the captured image as input and routes it into one child node. The leaf node fits a distribution over a set of 3D points in the world space that are projected from the training images using the ground truth camera pose and calibration parameters. Therefore, when evaluating a test frame with a decision tree, by routing an input point from root node to leaf node, a 3D-3D point correspondence can be easily established and further used for camera pose optimization. 
    
    \vspace{-1mm}
	\subsection{Outlier-aware Neural Tree} \label{sec:tree}
	\vspace{-1mm}
	\subsubsection{Hierarchical Space Partition for Decision Tree} \label{sec:tree_hsp}
    \vspace{-1mm}
    

    For the existing decision trees \cite{shotton2013scene,massiceti2017random,meng2017backtracking,meng2018exploiting,cavallari2017fly,cavallari2019real} developed for the camera relocalization problem, as there is no ground truth label for supervised training, the decision tree becomes ultimately a clustering strategy for the training data as observed in previous works \cite{cavallari2019real,castin2018clustering}. The decision rules at split nodes are learned via CLUS algorithm \cite{blockeel2000top} that uses variance reduction as the split criterion and achieves local-optimal hard data partition.
    In this paper, we propose to perform a hierarchical space partition for the target scene environment to construct our decision tree. We represent the entire scene as the root node, and iteratively partition the scene until reaching predefined depth. Each split node is responsible for a geometric region in the scene, and partitions this region into sub-regions of equal size for its child nodes. Each leaf node contains a set of 3D world coordinates in its covered local geometric region.
    The space partition strategy is illustrated in Figure \ref{figure:framework} and detailed below.
    
    
    Given a 3D scene model constructed in world space using the training sequence of <RGB-D image, camera pose>, we build an $m$-way decision tree, where $m$ is the $z$th power of 2. To perform a hierarchical space partition, we start from the root node which represents the entire scene environment. Then we compute the tight bounding box of the scene in the world coordinate system. We conduct iterative $z$ cuts to divide the bounding box into $m$ sub-boxes of equal volume size. In order to avoid the corner cases, such as long and narrow sub-boxes which may create challenges to learn the routing function, the decision rule is designed to encourage the divided bounding box to be similar to a cube. Specifically, to perform one cut on the bounding box of size ($w,h,l$), we find the longest edge over ($w,h,l$) and divide the box into two identical halves from the middle point of the edge. We iterate over such a process on the divided box until $z$ cuts are achieved. We perform such a top-down data partition iteratively for the nodes among all the levels. 
    
    The decision tree constructed in this way features several properties:
    (a) our constructed tree structure relies on the explicit space partition over the \textbf{3D scene environment} in the world space, not on the data partition of the visual observations (RGB+D) in the 2D camera space, then it requires the routing function to have more 3D understanding ability;
    (b) \textbf{each split node is physically meaningful}, and covers a specific geometric region, which is spatially related to other father or child nodes;
    (c) \textbf{the decision rules are predefined} by the $z$-cut space partition strategy introduced in the above paragraph and stay constant for all the nodes, instead of optimized via greedy algorithms to behave differently for different nodes;
    (d) the decision tree is more tolerant to \textbf{an $m$-way tree implementation}, not limited to a standard binary decision tree;
    (e) \textbf{the constructed tree structure is scene-dependent}, and may contain empty nodes that cover no geometric regions in the scene.
    Overall, the constructed decision tree via hierarchical space partition is more flexible in structure and physically meaningful compared to a standard decision tree in previous works.
    

    \vspace{-2mm}
    \subsubsection{Outlier-aware Neural Routing Function} \label{sec:route_func}
    \vspace{-2mm}

    Given an input sample, the purpose of the routing function at each split node is to send it to one of the child nodes. In our problem setting, the input sample is from the observed 2D RGB-D frame, and its ground truth label is determined by its corresponding location in the 3D world space. For purpose of accurate prediction, the routing function needs to understand the 3D scene context from 2D observations. Therefore, inspired by many previous works regarding point cloud classification \cite{qi2017pointnet,qi2017pointnet++} and point generation from 2D images \cite{fan2017point}, we take advantage of the point cloud processing framework to implement a neural routing function. We introduce the formulation of the input and network structure in detail below.
    
    \textbf{Input representation and sampling. }
    The input to the neural routing function is a query point $x_q$ that needs to be localized in the 3D world space, along with a set of context points $\{x_{o_i}\}_{i=1}^N$ in the neighbourhood of the query point. The input point is associated with color and depth, which are however both highly viewpoint dependent. In order to obtain better generalization ability in different viewpoints, given an input RGB-D frame, 
    We augment its depth channel via transforming it into the rotation-invariant geometric feature following PPF-FoldNet \cite{deng2018ppf}. To be specific, we firstly compute the oriented point cloud by projecting the full-frame depth into 3D camera space using camera calibration parameters, and calculating the pointwise normals in a 17-point neighbourhood \cite{hoppe1992surface}. Then we encode the query point and its context points into pair features,
    \begin{multline}
    \{ (x_q^{(p)},x_q^{(n)},x_{o_1}^{(p)},x_{o_1}^{(n)}),(x_q^{(p)},x_q^{(n)},x_{o_2}^{(p)},x_{o_2}^{(n)}),\cdots,\\
    (x_q^{(p)},x_q^{(n)},x_{o_N}^{(p)},x_{o_N}^{(n)}) \} \in \mathbb{R}^{12 \times N}
    \end{multline} 
    where $p$ and $n$ denotes the camera coordinate and normal, which form a 12-channel vector for each pair of oriented points ($x_q$, $x_{o_i}$).
    Each pair feature ($x_q^{(p)},x_q^{(n)},x_{o_i}^{(p)},x_{o_i}^{(n)}$) is then transformed into the rotation-invariant geometric representation \cite{deng2018ppf} that consists of three angles and one pair distance,
    \begin{multline}
    r = \{\angle(x_q^{(n)},x_q^{(p)}-x_{o_i}^{(p)}), \angle(x_{o_i}^{(n)},x_q^{(p)}-x_{o_i}^{(p)}),\\ \angle(x_q^{(n)},x_{o_i}^{(n)}), \Vert x_q^{(p)}-x_{o_i}^{(p)} \Vert_2\}  \in \mathbb{R}^{4}
    \end{multline} 
    
    Overall, for each input context point, it consists of both color $c$ and transformed rotation-invariant information $r$, represented as $\{x_{o_i}^{(c)},x_{o_i}^{(r)}\} \in \mathbb{R}^{7}$. Since the rotation-invariant feature for all context points is computed in the local reference frame with query point as origin, we omit the geometric feature and only take the color information as input for query point $x_q^{(c)} \in \mathbb{R}^{3}$.
    
    Given an input image, the query point for a split node is randomly sampled among the 2D image pixels whose projected 3D world coordinates belong to the split node. The context points are randomly sampled within the 3D neighbourhood ball of the query point. Ball query defines a radius, which is adaptively changed from level to level due to the varying size of covered 3D geometric region in our problem setting. In the implementation, we calculate the radius as the length of the longest edge of the covered bounding box.

    \textbf{Routing function design. }
    The routing function consists of two parts, the \textit{feature extraction} module and \textit{classification} module. The \textit{feature extraction} module leverages the pointwise multi-layer perception (MLP) to learn the features from both query point and context points inspired by the recent popular point cloud processing network PointNet \cite{qi2017pointnet}, while the \textit{classification} module combines the deep features from query point and context points to learn which child node the query point should be routed to.
    
    As the query point and context points are different in input channel, point number and impact for the classification task, we use different network parameters to encode their feature, specifically,
    \begin{equation}
        h_q = f_{featQ}(x_q^{(c)})
    \end{equation}
    \begin{equation}
        h_o = f_{featO}(\{x_{o_i}^{(c)},x_{o_i}^{(r)}\}_{i=1}^N)
    \end{equation}
    where $f_{featQ}$ and $f_{featO}$ are implemented with a 3-layer pointwise MLP (64-128-32/512), and extract the internal deep features ($h_q \in \mathbb{R}^{32}$, $h_o \in \mathbb{R}^{512}$) for query and context points respectively. $f_{featO}$ is followed with a max pooling layer to extract the global context feature. 
    
    Then $h_q$ and $h_o$ are concatenated and inputted into the classification module,
    \begin{equation}
        p = f_{class}(h_q,h_o)
    \end{equation}
    where $f_{class}$ is also implemented as a three-layer MLP (2048-1024-$k$), and outputs the probability ($p \in \mathbb{R}^{k}$) for all the child nodes. Since the constructed tree structure is scene dependent as mentioned in Sec. \ref{sec:tree_hsp}, the number of child nodes $k$ is adaptively changed from node to node and from scene to scene. As for supervision, we apply a cross entropy loss between the predicted probability $p$ and the ground truth label $y$ for supervision,
    \begin{equation}
        \mathcal{L} = -\sum_{i=1}^{k} \mathbbm{1}\{y_i = i\} \log\frac{\exp(p_i)}{\sum_{i=1}^{k} \exp(p_i)}
    \end{equation}
    where $y_i$ is the label for the $i$th child node.
    
    \textbf{Outlier rejection. }
    The aforementioned neural routing function is designed to route the input sample into one of the child nodes that are bound to 3D geometric regions. Given a query point belonging to dynamic regions in the test frame, the hierarchical routing functions will send it into one of the leaf nodes that contain the 3D world coordinates only from the static training scene, and it may establish an inaccurate 3D-3D correspondence for camera pose optimization. In order to solve this issue, we propose to reject the query points from dynamic regions as outlier, hence the established correspondence will be maintained in the confident static region.
    
    In order to achieve this goal, we further improve the neural routing function to output the probability vector $p$ of $k+1$ channels, where the additional channel refers to the outlier class. To generate the training samples for each split node from a given input image, the routing function considers any image pixel belonging to the current split node as \textit{inlier input} query point, which should be routed into child nodes. As the dynamic points in test environments are highly unpredictable, irregular, and do not exist in the training data, we simply consider any image pixel not covered by the current split node as \textit{outlier input} query point, which simulates the dynamic points and should be rejected without further routing. To train the routing function for each split node, the inlier and outlier input query points are sampled to be 3:1. Notice the outlier rejection strategy is incorporated into the neural routing function from the second level, since for the root node, all the image pixels belong to the inlier input.

    \vspace{-1mm}
    \subsubsection{HyperNetwork for the Routing Functions}
    \vspace{-1mm}
    
    In order to construct a $t$-level $m$-way tree, there are at most $m^{j-1}$ neural routing functions at level $j$ except for the bottom level that contains leaf nodes, and totally at most $\frac{m^{t-1}-1}{m-1}$ routing functions for the whole tree. It is both time-consuming and storage-inefficient to train so many deep networks. In order to decrease the training time and storage space for efficient deep learning, the previous works \cite{ha2017hypernetworks,fan2018decouple} unify the learnable parameters among different convolution layers in a network, time steps in a RNN, or hyper-parameters in an image filter within a standalone network, mostly known as HyperNetwork. More recent work \cite{fan2019general} further discovers that learning the normalization parameters with the HyperNetwork has similar performance as learning the convolution parameters, while the former case is more storage and running time friendly due to much less learnable parameters in the normalization layer compared to convolution layer.
    
    Inspired by these previous works, in this paper, we propose to leverage HyperNetwork to learn a single neural routing function for all the split nodes in the same level of a decision tree. Specifically, given the one-hot value $x_{node}$ that indicates the split node index, we learn to predict the learnable scale $\theta_{scale}$ and shift $\theta_{shift}$ parameters in the normalization layer of the \textit{classification} module in the neural routing function,
    \begin{equation}
        \theta_{scale},\theta_{shift} = f_{hyper}(x_{node})
    \end{equation}
    where $f_{hyper}$ refers to the HyperNetwork, and is implemented a three-layer MLP. The size of $\theta_{scale}$ and $\theta_{shift}$ depends on the channel number in the normalization layer. Then the normalization parameters in the \textit{classification} module is replaced with the predicted ones from the HyperNetwork,
	\begin{equation}
        p = f_{class}(h_q,h_o;\theta_{scale},\theta_{shift})
    \end{equation}
    Therefore, for a $t$-level tree, we only need to learn totally $t$ neural routing functions.
    
            \begin{table*}[t]
    \centering
        \begin{tabular}{lcccccc}
            \toprule
            \textbf{Method} & 
            \textbf{Score} $\uparrow$ & 
            \textbf{DCRE}$\mathbf{(0.05)}$ $\uparrow$&
            \textbf{DCRE}$\mathbf{(0.15)}$ $\uparrow$&
            \textbf{Pose}$\mathbf{(0.05m, 5^{\circ})}$ $\uparrow$& 
            \textbf{Outlier}$\mathbf{(0.5)}$ $\downarrow$&
            \textbf{N/A}\\
            \midrule
            HFNet \cite{sarlin2019coarse} & 0.373 & 0.064 & 0.103 & 0.018 & 0.360 & 0.000 \\
            HF-Net Trained \cite{sarlin2019coarse} & 0.789 & 0.192 & 0.300 & 0.073 & 0.403 & 0.000 \\
            NetVLAD \cite{arandjelovic2016netvlad} & 0.575 & 0.007 & 0.137 & 0.000 & 0.431 & 0.000 \\
            DenseVLAD \cite{torii201524} & 0.507 & 0.008 & 0.136 & 0.000 & 0.501 & 0.006\\
            Active Search \cite{sattler2016efficient} & 1.166  & 0.185 & 0.250 & 0.070 & \color{red}{0.019} & 0.690\\
            Grove \cite{cavallari2017fly} & 1.240 & 0.342 & 0.392 & 0.230 & 0.102 & 0.452 \\
            Grove V2 \cite{cavallari2019real} & 1.162 & \color{blue}{0.416} & 0.488 & \color{blue}{0.274} & 0.254 & 0.162 \\
            D2Net \cite{dusmanu2019d2} & \color{blue}{1.247} & 0.392 & \color{blue}{0.521} & 0.155 & 0.144 & 0.014 \\
            \midrule
            NeuralRouting (Ours) & \color{red}{1.441} & \color{red}{0.538} & \color{red}{0.615} & \color{red}{0.358} & \color{blue}{0.097} & 0.227\\
            \bottomrule
        \end{tabular}
        \vspace{2mm}
        \caption{Comparison on the test split of the RIO-10 benchmark w.r.t. the average score (1 + DCRE (0.05) - Outlier (0.5)), DCRE errors, camera pose accuracy and outlier ratio. \textit{N/A} denotes invalid/missing predictions. The red and blue numbers rank the first and second for each metric. }
    \label{table:relocalization_RIO}
    \vspace{-3mm}
    \end{table*}
    
    \begin{table*}[t]
    \centering
    \begin{tabular}{l|ccccccccc} 
        \toprule
        \textbf{Pose}$\mathbf{(0.05m, 5^{\circ})}$ $\uparrow$ & Chess & Fire & Heads & Office & Pumpkin & Kitchen & Stairs & Average \\
        \midrule
        Shotton et al. \cite{shotton2013scene} & 92.60\% & 82.90\% & 49.40\% & 74.90\% & 73.70\% & 71.80\% & 27.80\% & 67.60\% \\
        Guzman-Rivera et al. \cite{guzman2014multi} & 96.00\% & 90.00\% & 56.00\% & 92.00\% & 80.00\% & 86.00\% & 55.00\% & 79.30\% \\
        Valentin et al. \cite{valentin2015exploiting} & 99.40\% & 94.60\% & 95.90\% & 97.00\% & 85.10\% & 89.30\% & 63.40\% & 89.50\% \\
        Brachmann et al. \cite{brachmann2016uncertainty} & 99.60\% & 94.00\% & 89.30\% & 93.40\% & 77.60\% & \color{blue}{91.10\%} & 71.70\% & 88.10\% \\
        Schmidt et al. \cite{schmidt2016self} & 97.75\% & 96.55\% & \color{blue}{99.80\%} & 97.20\% & 81.40\% & \color{red}{93.40\%} & 77.70\% & 92.00\% \\
        Grove \cite{cavallari2017fly} & 99.40\% & 99.00\% & \color{red}{100.00\%} & 98.20\% & \color{red}{91.20}\% & 87.00\% & 35.00\% & 87.10\% \\
        Grove V2 \cite{cavallari2019real} & \color{red}{99.95}\% & \color{blue}{99.70\%} & \color{red}{100.00}\% & \color{blue}{99.48\%} & \color{blue}{90.85\%} & 90.68\% & \color{red}{94.20}\% & \color{red}{96.41\%} \\
        \midrule
        NeuralRouting (Ours) & \color{blue}{99.85\%} & \color{red}{100.00\%} & \color{red}{100.00\%} & \color{red}{99.80\%} & 88.80\% & 90.96\% & \color{blue}{84.20\%} & \color{blue}{94.80\%} \\
        \bottomrule
    \end{tabular}
    \vspace{2mm}
    \caption{Comparison on the 7-scenes dataset w.r.t. the camera pose accuracy. The red and blue numbers rank the first and second for each scene.}
    \label{table:relocalization_7scenes}
    \vspace{-4mm}
    \end{table*}
    
    \vspace{-1mm}
	\subsection{Camera Pose Estimation} \label{sec:pose}
	\vspace{-1mm}

	The core of our algorithm is a decision tree, which is the same as many previous camera relocalization works \cite{cavallari2017fly,cavallari2019real}. Therefore, we inherit similar optimization and refinement steps following \cite{cavallari2019real} for camera pose computation, which are introduced below.
	In order to generate the camera pose in $\mathbf{SE}(3)$, we firstly fit modes in the leaf nodes and then optimize the pose by leveraging the established 3D-3D correspondences. Each leaf node covers a set of 3D points (XYZ+RGB) in the world space projected from the 2D image pixels captured in the training sequence. Following \cite{valentin2015exploiting}, we detect the modes of the empirical distribution in each leaf node via mean shift \cite{comaniciu2002mean}, and then construct a Gaussian Mixture Model (GMM) via iteratively estimating a 3D Gaussian distribution for each mode. 
	After mode fitting of the leaf nodes, we leverage the preemptive locally-optimized RANSAC \cite{chum2003locally} for camera pose optimization. We start by generating 1024 pose hypotheses, each of which is computed by applying the Kabsch algorithm \cite{kabsch1976solution} on three randomly sampled 3D-3D point correspondences that relate the camera and world space. Given an observed point in camera space, its corresponding world coordinate is sampled from one random mode in the fitted GMM of the routed leaf node. We filter out the hypotheses that do not conform to the rigid body transformations following \cite{cavallari2017fly}, and regenerate the alternatives until they satisfy the above requirement. 
	The final camera pose is selected by iteratively re-evaluate and re-rank the hypotheses using the Mahalanobis distance, and discard the worse half until only one pose hypothesis is left.
	
	\textbf{Multi-leaves.}
	Given an input query point, the aforementioned pose optimization process fits modes only from the routed leaf node, which is common for the existing decision tree implementations as their routing function performs hard data partition and hence the input point can only be routed into a single leaf node. In contrast, the proposed neural routing function performs a ``soft'' data partition with predicted probability $p$, hence the input point can be ``routed'' to all the leaf nodes with different accumulated probabilities through probability multiplication of all routed split nodes. Motivated by the above observation, to achieve more robust pose optimization, we fit the mode by combining the world coordinates from multiple routed leaf nodes with highest accumulated probabilities, instead of a single leaf node. In the implementation, we use four leaf nodes, which works the best experimentally, for mode fitting.
	
	\textbf{Pose refinement.}
	Last but not least, we follow \cite{cavallari2019real} to incorporate our camera relocalizer into a 3D reconstruction pipeline for further camera pose refinement, which mainly consists of ICP \cite{besl1992method} and model-based hypothesis ranking.

	\section{Experiments}
	\subsection{Implementation Details}
	
	\textbf{Tree structure.} For all the experiments in this paper, we implement the 5-level 16-way tree for scene partition, thus a perfect tree structure consists of 4369 nodes in this case. During our implementation, according to the specific scene geometry, such a tree contains about 2000 to 3000 valid nodes. 
	
	\textbf{Training details.} The neural routing functions are implemented in PyTorch. Benefited from the design of HyperNetwork, we only train 5 neural routing functions. Each routing function is trained for 60 epochs with a batch size of 256. The network weights are optimized with Adam \cite{kingma2014adam} whose initial learning rate is 0.001 and betas are (0.9,0.999). The initial learning rate is halved every 20 epochs until the end. The number of context points is set as 600 all the time.
	
	\subsection{Dataset}
	
	We test our proposed algorithm on two camera relocalization benchmarks, RIO-10 \cite{wald2020beyond} and 7-scenes \cite{shotton2013scene}, which are developed for dynamic and static indoor scenes respectively. The RIO-10 dataset includes 10 real indoor environments, each of which is scanned several times over different time periods, and demonstrates the common geometric and illumination changes in dynamic environments. This dataset is separated into training/validation/test split, while the test results should be obtained via submission to their online benchmark. The 7-scenes dataset contains only training and test set, and is the most popular camera relocalization benchmark for the static indoor environments in the past.
	
	\setlength{\tabcolsep}{6pt}
    \begin{table}[t]
    \centering
    \begin{tabular}{c|c} 
        \toprule
        & \textbf{Pose}$\mathbf{(0.05m, 5^{\circ})}$ $\uparrow$ \\
        \midrule 
        Ours w/o outlier labels & 25.14\%\\
        Ours w/o multi-leaves & 25.80\%\\
        5-level 8-way Tree & 24.60\%\\
        3-level 16-way Tree & 16.75\%\\
        4-level 16-way Tree & 25.31\%\\
        \midrule 
        Ours (5-level 16-way Tree) & 27.05\%\\
        Ours w. pose refinement & 31.99\%\\
        \bottomrule
    \end{tabular}
    \vspace{2mm}
    \caption{Ablation study on the validation set (10 scenes) of the RIO-10 benchmark. Ours is the full pipeline of our algorithm. }
    \label{table:ablation}
    \end{table}
    
    	\begin{figure*}
		\centering
		\includegraphics[width=0.94\linewidth]{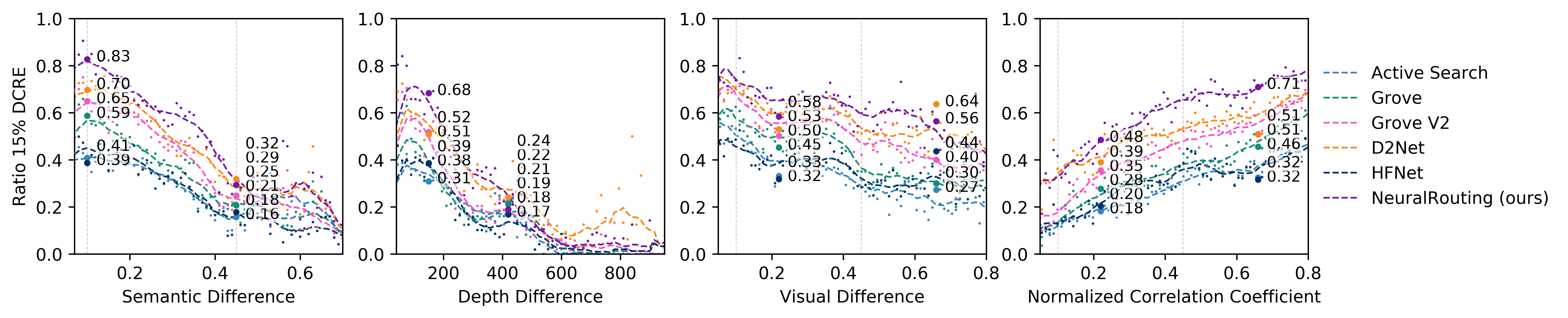}
		\caption{The charts show the performance (\textbf{DCRE}$\mathbf{(0.15)}$) of compared approaches with respect to semantic (semantic difference), geometric (depth difference) and visual change (NSSD, Normalized Correlation Coefficient) as introduced in RIO-10 dataset \cite{wald2020beyond}.}
		\label{figure:curve}
	\end{figure*}
	
	\subsection{Evaluation Metrics}
	
	In order to evaluate the quality of estimated camera pose, we adopt the common camera pose accuracy \textbf{Pose}$\mathbf{(0.05m, 5^{\circ})}$, which is computed as the proportion of test frames whose translation error is within 5 centimeters and angular error is within 5 degrees. In the RIO-10 benchmark \cite{wald2020beyond}, we further adopt their proposed new metric \textbf{DCRE}, short for Dense Correspondence Re-Projection Error, which is computed as the average magnitude of the 2D correspondence displacement normalized by the image diagonal. The displacement is calculated between 2D projections of the underlying 3D model using the ground truth and predicted camera poses. DCRE depicts an error that correlates with the visual perception, not only with the absolute camera pose. Then \textbf{DCRE}$\mathbf{(0.05)}$ and \textbf{DCRE}$\mathbf{(0.15)}$ are the percentage of test frames whose DCRE is within 0.05 or 0.15, while \textbf{Outlier}$\mathbf{(0.5)}$ describes the opposite case, which is the percentage of test frames whose DCRE is above 0.5. 
	    
    \begin{figure}
		\centering
		\includegraphics[width=0.94\linewidth]{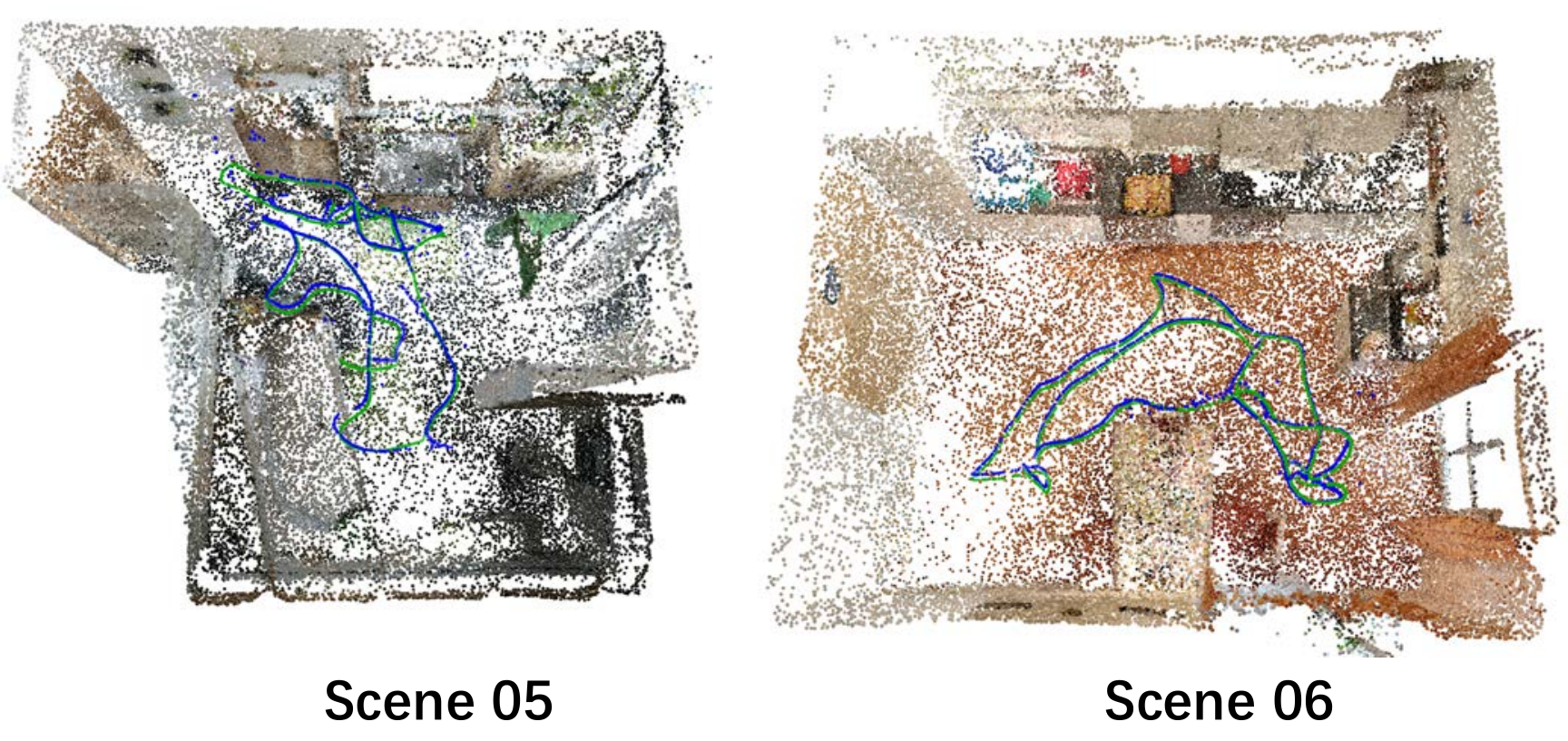}
		\caption{Ground truth (green) and predicted (blue) camera pose trajectory on the validation set of RIO-10 dataset.}		
		\label{figure:trajectory}
  		\vspace{-3mm}
	\end{figure}

	\subsection{Numerical Results}
    
    We compare our algorithm with all the other approaches on the test split of the RIO-10 dataset, shown in Table \ref{table:relocalization_RIO}. Among all the metrics that evaluate the quality of camera pose estimations, our algorithm ranks the first except for \textbf{Outlier}$\mathbf{(0.5)}$, where our performance is the second best. Regarding the camera pose accuracy \textbf{Pose}$\mathbf{(0.05m, 5^{\circ})}$, which is more common and directly measures the pose quality, our result (0.358) surpasses the state-of-the-art approaches (0.274) significantly by about 30\%. It demonstrates the effectiveness and robustness of our proposed outlier-aware neural tree on the dynamic indoor environments. 
    
    We further test our algorithm on the popular camera relocalization benchmark 7-scenes for static indoor scenes, shown in Table \ref{table:relocalization_7scenes}. Our algorithm ranks the second place on the averaged camera pose accuracy among all the existing approaches, and lags behind the best performance within a very small gap. It further shows the excellent generalization ability of our algorithm on static scenes, though it is developed for dynamic environments. Note the baseline results in RIO-10 and 7-scenes datasets are from the official online benchmark and the recent SOTA relocalization paper \cite{cavallari2019real}, respectively.
    
    We test the running time of our algorithm on GPU. For a single image, the camera pose optimization and refinement take around 100 ms and 150 ms separately, similar to the previous decision tree based approach \cite{cavallari2019real}. The neural routing runs for 480 ms, while its light version without considering multiple leaves during routing takes only 100 ms yet achieves similar camera pose accuracy as verified in Table \ref{table:ablation}.


	\subsection{Ablation Study}
	
	To justify the effectiveness of our algorithm design, we conduct an ablation study as shown in Table \ref{table:ablation}, which is evaluated on the validation set of RIO-10 dataset. Space partition is important for our neural tree construction, hence we firstly test different strategies to partition the scene by varying the hyper-parameters $t$ and $m$ in the $t$-level $m$-way tree. We observe that as the number of levels $t$ decreases, the camera pose accuracy also degrades, this is mainly due to the increased box size in leaf node, which creates difficulty in fitting a good distribution and sampling effective world coordinates. 
	Notice the leaf nodes in 4-level 16-way tree and 5-level 8-way tree have the same box size, while 16-way tree is better in camera pose accuracy. This is mainly because the 4-level tree has fewer routing functions to be trained, and hence accumulates less error from the deep network during hierarchical routing.
	Finally we validate the design of outlier classification and multi-leaves in camera pose optimization by removing them from the entire framework, and observe worse pose accuracy as expected.

	\subsection{Analysis}
	
	\textbf{Pose trajectory.} We visualize the pose trajectory of both our estimations and ground truth on two scenes in the validation split of RIO-10 dataset in Figure \ref{figure:trajectory}. We observe a significant overlap between the two trajectories, which verifies the effectiveness of our algorithm in dynamic indoor environments.
	
	
	\textbf{Performance against various scene changes.} To discover how our algorithm is robust to scene changes compared to other approaches, we visualize the overall performance of each method with images of increasing visual, geometric and semantic change as defined in RIO-10 dataset, in Figure \ref{figure:curve}. We are glad to see that our plotted curve is almost the best among all the different types of scene changes. It further verifies our algorithm for camera relocalization in dynamic indoor scenes.


	\section{Conclusion}
	In this paper, we propose a novel outlier-aware neural tree to achieve accurate camera relocalization in dynamic indoor environments. Our core idea is to construct a decision tree via hierarchical space partition of the scene environment, and learn a neural routing function to reject the dynamic input points during the level-wise routing process. Extending our work to only the RGB input and generalization to novel environments are more realistic yet challenging settings, which are treated as valuable future directions to explore.
	
	\section*{Acknowledgements}
	This work was supported in part by National Key R\&D Program of China (2019YFF0302902), National Science Foundation of China General Program grant No. 61772317, NSF grant IIS-1763268, a grant from the Samsung GRO program, a Vannevar Bush Faculty Fellowship, and a gift from Amazon AWS ML program.

	\section{Appendix}
	The appendix provides the additional supplemental material that cannot be included into the main paper due to its page limit:
	
	\begin{itemize}
		\item More Space Partition Strategies.
		\item Extension to Neural Forest.
		\item Unified Neural Routing Function -- PointNet++.
		\item Ablation for HyperNetwork.
	\end{itemize} 
	
	\subsection*{A. More Space Partition Strategies} \label{sec:space}
	
	In the main paper, we evaluate different space partition strategies by varying the hyper-parameters in a $t$-level $m$-way tree. In this section, we conduct more experiments by constructing the covered bounding boxes in different manners, which is another important factor that influences the tree structure. To be specific, we firstly follow the axis of world coordinate system and compute the axis-aligned minimum bounding box (AABB) of the entire scene as the root node for space partition. This is our \textbf{original} implementation in the main paper. To explore more space partition strategies, we rotate the scene model along the x and y axis by \textbf{30$^{\circ}$}, and calculate the new AABB. Similarly, we also obtain a version by rotating \textbf{60$^{\circ}$}. The bounding boxes constructed above all follow the world coordinates, and may leave many blank 3D spaces in the box, which does not make full use of the neural routing functions. To resolve this issue, we obtain the \textbf{compact} box by recalculating the world coordinate system of the scene using PCA \cite{pearson1901liii}, and fit the tightest bounding box along the new axis. In order to alleviate the potential influence of coordinate axis to camera pose optimization as observed in \cite{brachmann2018learning,brachmann2017dsac}, we further rotate the compact box to align with the default world coordinate axis for a fair comparison with other boxes.
	
\begin{figure}[t]
		\centering
		\includegraphics[width=0.94\linewidth]{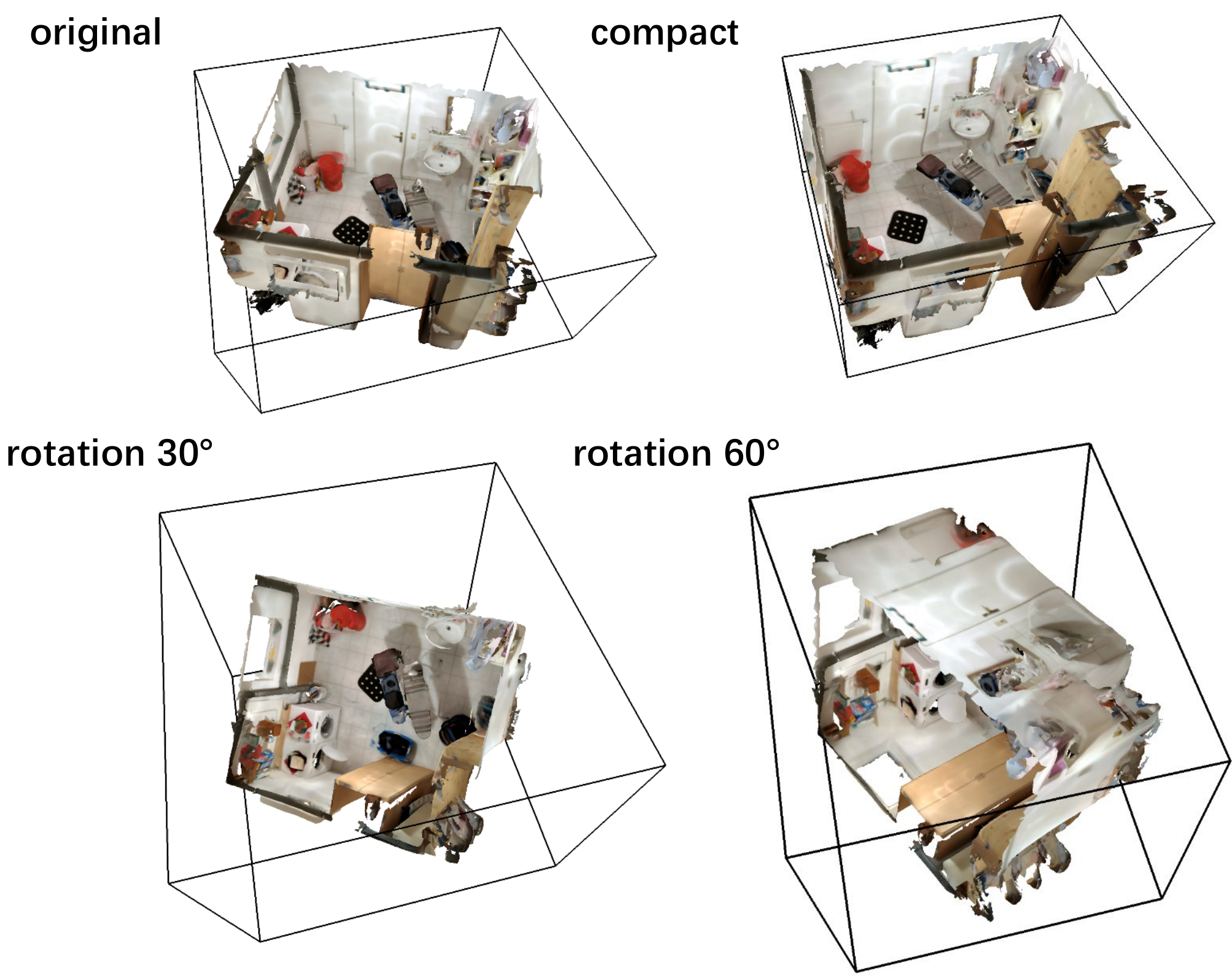}
		\caption{Different space partition strategies via bounding box construction.}		\label{figure:box}
		\vspace{-5mm}
	\end{figure}

	We illustrate the different bounding box constructions in Figure \ref{figure:box}. We observe that regarding the compactness between the bounding box and 3D scene model, compact box > original box > rotation 60$^{\circ}$ > rotation 30$^{\circ}$.
	The corresponding numerical results of the above space partition strategies are shown in Table \ref{table:result}. Consistent with the compactness, the camera pose accuracy also follows the same order: compact box > original box > rotation 60$^{\circ}$ > rotation 30$^{\circ}$. It indicates an interesting observation that the more compact the box is, the higher the pose accuracy can be achieved by our algorithm. This is mainly because in a compact box, the geometric regions are more uniformly sampled among all the split nodes in the decision tree, which strengthens the utilization of the neural routing functions.
	
	\setlength{\tabcolsep}{6pt}
	\begin{table}[t]
		\centering
		\begin{tabular}{c|c} 
			\toprule
			& \textbf{Pose}$\mathbf{(0.05m, 5^{\circ})}$ $\uparrow$ \\
			\midrule 
			original + rotation 30$^{\circ}$ & 51.05\% \\
			original + rotation 60$^{\circ}$ & 52.98\% \\
			original box & 54.93\% \\
			compact box & 56.68\% \\
			\midrule 
			forest & 58.29\% \\
			\midrule 
			PointNet++ & 1.93\% \\
			\bottomrule
		\end{tabular}
		\vspace{2mm}
		\caption{Camera pose accuracy on the scene 01 in the validation set of RIO-10 dataset.}
		\label{table:result}
		\vspace{-4mm}
	\end{table}
	
	\subsection*{B. Extension to Neural Forest}
	
	In the existing camera relocalization works implemented with decision trees \cite{cavallari2017fly,cavallari2019real}, they usually train a number of trees on the same scene to obtain a forest for the pose optimization. In this way, the final prediction of the forest is simply the union of the fitted modes among all the trees. In these works, the decision rule for each split node is adaptively learned as either color or depth comparison from the training data. Hence by simply sampling different input samples in the same scene, they are able to learn different decision trees.
	
	Motivated by the previous works, we further extend our proposed neural tree to neural forest by training multiple trees. However, in our work, the decision rules are predetermined by the space partition strategy. In order to enable the diversity of multiple trees, we adopt the four different space partition strategies introduced in Section \ref{sec:space}, and unify their predictions to form a neural forest. The numerical results are shown in Table \ref{table:result}. We are glad to observe that the performance can be further upgraded by the utilization of a forest.
	
	\subsection*{C. Unified Neural Routing Function -- PointNet++}
	
	The main paper employs the hierarchical node-wise neural routing functions to classify each input query point into one of the leaf nodes. This can be naturally viewed as the point-wise segmentation task, where each segmentation label refers to one leaf node. As the input is formulated as the form of point cloud, we can achieve a unified neural routing function by directly leveraging the popular state-of-the-art point cloud segmentation network PointNet++ \cite{qi2017pointnet++}. In our problem setting, the PointNet++ takes the colored point cloud from a single frame as input, and outputs the point-wise segmentation mask. In this case, each input point is the query point and also serves as the context point for the other query points. In this unified neural routing function, the outlier rejection is not an option anymore and excluded from the segmentation label. We adopt the MSG as the PointNet++ backbone in the implementation.
	
	Its numerical result is shown in Table \ref{table:result}, which performs much worse compared to our neural routing function implementation. It demonstrates the effectiveness of our unique neural tree design.
	
	\subsection*{D. Ablation for HyperNetwork}
	
	In our algorithm, HyperNetwork unifies the network parameters of all the neural routing functions from the same level into a single network, and hence saves much storage space and training time. However, as observed in the previous work \cite{fan2019general}, HyperNetwork may potentially degrade the performance compared to the version that separately trains each network. To investigate the potential influence of HyperNetwork on the performance of neural routing functions, we select three split nodes for each level in our neural tree, and compare their classification accuracy between the implementations with and without HyperNetwork as shown in Table \ref{table:hypernetwork}.
	Interestingly, we observe that the usage of HyperNetwork only degrades the performance within a reasonable range similar to the past experience \cite{fan2019general}.
	
	\setlength{\tabcolsep}{3pt}
	\begin{table}[t]
		\centering
		\begin{tabular}{c|ccc|ccc} 
			\toprule
			& \multicolumn{3}{c|}{w. HyperNet} & \multicolumn{3}{c}{w/o HyperNet} \\
			\midrule 
			& level 2 & level 3 & level 4 & level 2 & level 3 & level 4 \\
			\midrule 
			node 1 & 58.1\% & 75.9\% & 70.5\% & 60.5\% & 79.5\% & 75.0\% \\
			node 2 & 57.9\% & 66.5\% & 69.5\% & 56.8\% & 62.6\% & 69.3\% \\
			node 3 & 23.6\% & 48.5\% & 49.3\% & 28.9\% & 66.7\% & 50.0\% \\
			\midrule 
			average & 46.5\% & 63.63\% & 63.1\% & 48.7\% & 69.9\% & 64.7\% \\
			\bottomrule
		\end{tabular}
		\vspace{2mm}
		\caption{Ablation study of HyperNetwork on the scene 01 in the validation set of RIO-10 dataset. For each level, we collect three split nodes to evaluate their classification accuracy on the validation set.}
		\label{table:hypernetwork}
		\vspace{-4mm}
	\end{table}

{\small
\bibliographystyle{ieee_fullname}
\bibliography{egbib}
}

\end{document}